\documentclass[journal]{IEEEtran}
\IEEEoverridecommandlockouts
\ifCLASSOPTIONcompsoc
\usepackage{rotating}  
  \usepackage[utf8]{in an  pu withtenc}
 \usepackage{multirow} 
 \usepackage{booktabs} % For formal tables
\usepackage{longtable} % For tables that span mul aa aapko phone
.
tiple pages
\usepackage{hyperref} % For hyperlinks
\usepackage{lscape} % If you need to landscape the table
\usepackage[nocompress]{cite}
..
\else
  \usepackage{cite}
\fi

\usepackage{amsmath}
\usepackage{amssymb}
\usepackage{amsfonts}
\usepackage{tikz}

\usepackage{graphicx}
\usepackage{textcomp}
\usepackage{xcolor}
\usepackage{multirow}
\usepackage{soul}
\usepackage{algorithm}
\usepackage{algpseudocode}
\usepackage{colortbl}
\usepackage{bm}
\usepackage{float}
\usepackage{CJK}
\usepackage{caption}
\usepackage{subcaption}
\usepackage{algorithm}
\usepackage{algorithmicx}

\usetikzlibrary{automata, arrows.meta, positioning}

\usepackage{diagbox}

\newcommand\encircle[1]{%
\tikz[baseline=(X.base)] 
  \node (X) [draw, scale=0.75, shape=circle, inner sep=0, fill=black, text=white, minimum size=0em] {\strut #1};}

\newcommand{\cmmnt}[1]{}  

\def\BibTeX{{\rm B\kern-.05em{\sc i\kern-.025em b}\kern-.08em
    T\kern-.1667em\lower.7ex\hbox{E}\kern-.125emX}}

\ifCLASSINFOpdf
\else
\fi

\usepackage{todonotes}

\usepackage[a4paper, total={184mm,239mm}]{geometry}
\def\BibTeX{{\rm B\kern-.05em{\sc i\kern-.025em b}\kern-.08em
    T\kern-.1667em\lower.7ex\hbox{E}\kern-.125emX}}

\begin{document}
\newcommand{\algrule}[1][.2pt]{\par\vskip.5\baselineskip\hrule height #1\par\vskip.5\baselineskip}

% \title{\huge IA\textsuperscript{2}: \ul{I}ntermittent-\ul{A}ware \ul{I}nference \ul{A}rchitecture Realizing Batteryless Intelligence IoTs\vspace{-0.4em}
% }
\title{\huge ATM-Net: \underline{A}daptive \underline{T}ermination and \underline{M}ulti-Precision Neural \underline{Net}works for Energy-Harvested Edge Intelligence}

\author{Neeraj Solanki, Sepehr Tabrizchi, Samin Sohrabi, Jason Schmidt, Arman Roohi\\
{\small Department of Electrical and Computer Engineering, University of Illinois Chicago, Chicago, IL, USA}\\
\{nsola5,stabr,ssohra2,jschm25,aroohi\}@uic.edu \vspace{-2.8em}}

% \author{Sepehr Tabrizchi$^{\dagger}$, Omer Kurkutlu$^{\dagger}$,
% Shaahin Angizi$^{\ddagger}$, and Arman Roohi$^{\dagger}$^{*}  \vspace{0.3em}\\ \small 
% $^{*}$School of Computing, University of Nebraska-Lincoln, Lincoln, NE, USA\\
% $^\dagger$Department of Electrical and Computer Engineering, University of Illinois Chicago, IL, USA\\
% $^\ddagger$Department of Electrical and Computer Engineering, New Jersey Institute of Technology, Newark, NJ, USA\\
% % $^\S$Department of Computer Science, University of California Irvine, Irvine, CA, USA\\
% shaahin.angizi@njit.edu, aroohi@uic.edu \vspace{-2em}
% \\}

% \author{
%         Nedasadat~Taheri,~\IEEEmembership{Student Member,~IEEE,}
%         Sepehr~Tabrizchi,~\IEEEmembership{Student Member,~IEEE,}
%         Shaahin~Angizi,~\IEEEmembership{Senior Member,~IEEE,}
%         and~Arman~Roohi,~\IEEEmembership{Senior Member,~IEEE}\vspace{-1em}% <-this % stops a space
% \thanks{S. Tabrizchi, N. Taheri, and A. Roohi are with the School of Computing, University of Nebraska-Lincoln, Lincoln, NE, 68588 USA e-mail: (aroohi@unl.edu).}% <-this % stops a space
% \thanks{S. Angizi are with the Department of Electrical and Computer Engineering, New Jersey Institute of Technology (NJIT), Newark, NJ, USA e-mail: (shaahin.angizi@njit.edu).}
% }

\maketitle
% \vspace{-0.5em}
\begin{abstract}
ATM-Net is a novel neural network architecture tailored for energy-harvested IoT devices, integrating adaptive termination points with multi-precision computing. It dynamically adjusts computational precision (32/8/4-bit) and network depth based on energy availability via early exit points. An energy-aware task scheduler optimizes the energy-accuracy trade-off. Experiments on CIFAR-10, PlantVillage, and TissueMNIST show ATM-Net achieves up to 96.93\% accuracy while reducing power consumption by 87.5\% with Q4 quantization compared to 32-bit operations. The power-delay product improves from 13.6J to 0.141J for DenseNet-121 and from 10.3J to 0.106J for ResNet-18, demonstrating its suitability for energy-harvesting systems.
\end{abstract}

% \begin{IEEEkeywords}
%  Sensor attacks, processing-in-sensor, neural network, vision sensor 
% \vspace{-0.75em}
% \end{IEEEkeywords}

\section{Introduction}
The proliferation of Internet of Things (IoT) devices is reshaping our digital landscape at an unprecedented pace, with projections indicating over 75 billion connected devices by 2025. This explosive growth drives a fundamental transformation in how we process and analyze data, necessitating a shift from centralized cloud computing to distributed edge intelligence. Traditional cloud-centric architectures, while powerful, face mounting challenges, including network congestion, communication latency, privacy concerns, and escalating energy consumption in data centers. For instance, data centers currently consume approximately 1\% of global electricity, with this figure projected to reach 3-8\% by 2030 \cite{verma2022paradigm,liu2020energy,ericsson2024climate,cordis2024batteries}.

The migration toward edge intelligence, where data processing occurs directly on IoT devices, offers a promising solution to these challenges. Edge computing not only reduces network bandwidth consumption and latency but also enhances privacy by keeping sensitive data local. However, this paradigm shift introduces critical energy management challenges. Currently, most IoT devices rely on batteries as their primary power source, creating substantial environmental and logistical problems. The environmental impact is particularly concerning – it's estimated that over 3 billion batteries are discarded annually in the United States alone, with most containing toxic materials that pose significant environmental hazards. Moreover, the maintenance costs associated with battery replacement in large-scale IoT deployments are prohibitive, often exceeding the initial hardware costs over the device's lifetime.
These challenges have spurred interest in energy-harvested IoT devices powered entirely by energy harvesting. Such devices eliminate battery-related environmental concerns and maintenance requirements while offering theoretically indefinite operational lifetimes. However, operating sophisticated computational tasks, particularly neural network inference, on harvested energy presents unprecedented challenges. Energy harvesting is inherently unstable, with power availability fluctuating based on environmental conditions. Traditional neural networks, designed for stable computing environments, are ill-equipped for such intermittent operation scenarios.

Traditional convolutional neural networks (CNNs) process all inputs through every layer, regardless of the input's complexity. This one-size-fits-all approach is inefficient, particularly when handling simple inputs that could be accurately classified using fewer computational steps.
The inefficiency is especially problematic in energy harvesting systems that rely on harvested energy, such as solar-powered IoT sensors. These devices face extreme energy fluctuations - a solar-powered sensor might have 100 times more energy available at noon compared to dawn or dusk. Current neural network architectures aren't designed to adapt to such dramatic energy variations.
Multi-exit architectures offer a solution by adding decision points throughout the network\cite{teerapittayanon2016branchynet}. At each point, the system can evaluate if it has enough confidence to make a prediction, allowing it to exit early for simpler inputs rather than forcing all computations through the entire network. This adaptive approach reduces both processing time and energy usage, making it particularly valuable for devices that operate under strict power and time constraints.

This paper introduces a novel adaptive neural network architecture that synergistically combines multi-precision computing with multiple termination points, specifically designed for energy-harvested IoT devices. Unlike previous approaches, our system dynamically adjusts both computational precision and network depth based on instantaneous energy availability. This dual-adaptation mechanism enables fine-grained control over the energy-accuracy trade-off, allowing the system to maintain operational continuity under highly constrained and variable energy conditions.

% \section{Motivation}
% The growing demand for deploying deep learning models in resource-constrained environments, such as edge devices, IoT systems, and energy-harvesting systems (EHS), presents significant challenges due to the high computational and memory requirements of state-of-the-art architectures . 
% Conventional neural networks, which fully propagate all inputs through every layer, lack adaptability and result in inefficient use of resources, especially for simpler inputs that do not require deeper feature extraction.

% Additionally, quantization further enhances efficiency by reducing the precision of weights and activations, significantly lowering memory and computational overhead. 
\section{Proposed ATM-Net}
% \begin{itemize}
%     \item How to determine optimal exit point placement?
%     \item What precision levels to use for different layers?
%     \item How to handle state preservation during power failures?
%     \item What are the best confidence thresholds for early exits given varying energy availability?
% \end{itemize}

% \begin{enumerate}
%     \item Develop an adaptive neural network system that can:
%     \begin{itemize}
%         \item Dynamically adjust precision based on available energy
%         \item Exit early when confidence thresholds are met
%         \item Handle intermittent power scenarios typical in energy harvesting systems
%     \end{itemize}
%     % \item Key Technical Components:
%     % \begin{itemize}
%     %     \item \sepehr{Energy Management}
%     %     \item 1- Implement energy availability monitoring, 2- Create an energy-aware scheduler that decides (When to use lower precision computations; When to take early exits; When to suspend operations until more energy is available)
%     % \end{itemize}
%     \end{enumerate}

Adaptive neural networks are innovative architectures capable of dynamically modifying their computation paths to balance efficiency, accuracy, and resource usage. This adaptability makes them particularly well-suited for deployment in constrained environments, such as energy harvesting systems (EHS) and edge devices. This study focuses on two pivotal aspects of adaptive neural networks: \textit{quantization} and \textit{multi-termination}, both of which contribute to optimizing performance and energy efficiency while maintaining task-specific accuracy.
\begin{figure}[t!]
    \centering
    \includegraphics[width=\linewidth]{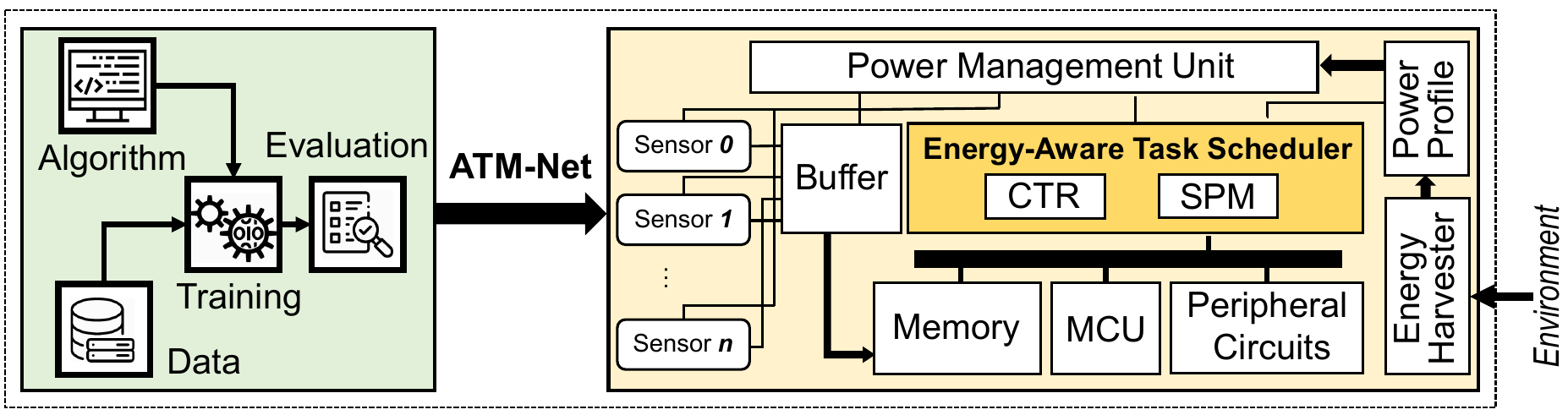}%\vspace{-0.5em}
    \caption{Proposed architecture, including ATM-Net and the energy-aware task scheduler.}
    \vspace{-1em}
    \label{fig:main}
\end{figure}

\subsection{Quantization Synergy}
Quantization is a pivotal optimization technique that reduces the precision of weights and activations in neural networks, significantly enhancing computational efficiency and energy consumption. Herein, we explored 8-bit (Q8) and 4-bit (Q4) quantization schemes to optimize neural network performance under various resource constraints. These methods were applied uniformly across all layers, leveraging Quantization-Aware Training (QAT)\cite{jacob2018quantization} to ensure accuracy retention. QAT simulates quantization effects during both the forward and backward passes, enabling the network to adapt its parameters and mitigate the loss of precision. The loss function in QAT is computed as:
\begin{equation}
L = \frac{1}{N} \sum_{i=1}^{N} \ell \left( y_i, f \left( \hat{W}, x_i \right) \right)
\end{equation}
where \(\hat{W}\) represents the quantized weights, \(f(\cdot)\) denotes the model’s output, \(\ell\) is the loss function (e.g., cross-entropy), and \(x_i, y_i\) are the input and ground truth for the \(i\)-th sample. This approach ensures that quantized models maintain robustness and achieve accuracy levels comparable to their full-precision counterparts.
The Q8 quantization scheme used a linear quantization approach, where weights \(W\) and activations \(A\) were quantized as:
\begin{equation}
\hat{W} = \text{clip} \left( \text{round} \left( \frac{W - \alpha}{\Delta} \right), \text{min}, \text{max} \right) \cdot \Delta + \alpha
\end{equation}
with \(\alpha\) as the zero-point and \(\Delta\) as the scale factor, clipping values to the range \([-128, 127]\) for 8Q representation. This method balances computational efficiency and accuracy, making it suitable for general-purpose deployment in moderate resource-constrained environments.
For Q4 quantization, a Custom Observer method was employed to address the reduced dynamic range of \([-8, 7]\). This observer dynamically adjusted the scale factor \(\Delta\) to minimize quantization errors, ensuring the preservation of critical features despite the lower bit-width. This approach proved effective in ultra-constrained environments, such as edge devices with strict energy budgets, by significantly reducing memory and computational demands while maintaining reasonable accuracy.

\subsection{Multi-Exit CNN Models}
Multi-termination is an advanced design strategy that introduces early exit (EE) points within neural network architectures, enabling dynamic termination of computations based on input complexity. By incorporating intermediate prediction layers, the network generates outputs earlier in the forward pass for inputs with high-confidence predictions. This approach optimizes computational efficiency and energy consumption while preserving competitive accuracy, making it particularly advantageous for resource-constrained environments and low-latency applications. EE points are strategically placed after layers capable of extracting essential features, such as low- and mid-level abstractions. Initial layers focus on fundamental patterns like edges and textures, making them ideal for shallow exits. Intermediate layers refine these features, processing moderately complex inputs, while deeper layers handle intricate representations, utilizing the network's full computational depth for challenging inputs. This hierarchical placement ensures simpler inputs exit early, significantly reducing computational overhead, while more complex inputs proceed to deeper layers for higher accuracy. To facilitate early termination, confidence scores are evaluated at each exit point using the softmax probability distribution:
\begin{equation}
C_i = \max(P(y|x; \theta_i)),
\end{equation}

where \(C_i\) represents the confidence score at the \(i\)-th exit, \(P(y|x; \theta_i)\) is the softmax probability of the predicted class, and \(\theta_i\) denotes the network parameters up to the \(i\)-th exit. If \(C_i\) exceeds a predefined threshold \(T_i\), the computation halts, and the corresponding prediction is outputted. These thresholds are calibrated through experimentation to balance computational savings with prediction accuracy. For datasets with high-dimensional or intricate features, deeper layers capture the required abstractions, while simpler datasets benefit from shallow exits. This adaptability underscores the scalability and efficiency of multi-termination, making it a robust approach for modern neural network applications.

\subsection{Target Networks}
The first model employed in our study is a modified ResNet-18\cite{resnet2015} architecture integrated with three exit mechanisms: Early Exit 1 (EE1), Early Exit 2 (EE2), and the Main Exit (ME). ResNet-18 architecture is designed for efficient feature extraction through its residual block design. In our adaptation, EEs are strategically placed after the first and second residual layers, leveraging intermediate feature representations for early predictions. EE1, located after the first residual layer, extracts low-level features such as edges and textures. This exit comprises an adaptive average pooling layer, which reduces spatial dimensions, followed by a fully connected layer tailored to the dataset’s classification requirements. EE1 allows the model to terminate computations dynamically for simpler inputs that do not require deeper feature processing, significantly reducing computational overhead. EE2 is positioned after the second residual layer and captures mid-level feature abstractions, such as object parts and more complex patterns. Similar to EE1, it consists of an adaptive average pooling layer and a fully connected layer, enabling moderately complex inputs to be classified effectively without traversing the entire network. The ME handles high-level feature abstractions at the final stages of the network, ensuring accurate predictions for complex inputs requiring comprehensive processing. This multi-termination strategy in ResNet-18, combined with Q8 and Q4 quantizations facilitated through QAT.

The second model employed is a modified DenseNet-121\cite{densenet2017} architecture, also incorporating three exits: EE1, EE2, and ME. DenseNet-121 is characterized by its dense connectivity, where each layer receives input from all preceding layers. This dense connectivity promotes efficient feature reuse, enhancing parameter efficiency and gradient flow during training. 
EE1 is integrated after the second dense block and is responsible for capturing low- and mid-level features. It consists of an adaptive average pooling layer and a fully connected layer, similar to the setup in ResNet-18. This exit provides predictions for inputs that are less complex, reducing the need for further computation. EE2, located after the third dense block, handles more intricate feature abstractions and serves inputs of moderate complexity. ME processes the most challenging inputs, utilizing DenseNet-121’s robust feature propagation capabilities to ensure accurate predictions.DenseNet-121’s architecture inherently complements the early exit strategy, as its dense connectivity ensures comprehensive feature extraction at all levels. 
This integration of three exits, alongside the application of QAT, further enhances the model’s efficiency for deployment on resource-constrained hardware, while maintaining high accuracy across a variety of datasets.

\subsection{Proposed Energy-Aware Task Scheduler (EATS)}
% \sepehr{Add a paragraph related batteryless (EHS) system and how they work}

% Adaptive scheduler consists of a controller (CTR) and smart power management (SPM) to change the operation's precision and finish the computation sooner using the designed early exits in case of power limitation.
% In batteryless edge devices, there are two critical points related to energy: first, the charging rate, and second, the remaining energy in the capacitor. To find the best performance, the SPM needs to monitor both parameters. In the proposed system precision of computation has more effect on the power consumption rate and also the gained accuracy. In addition, as mentioned, this precision cannot change during calculation for an input. As a result, SPM will measure the charging rate and, on the basis of the charging rate, decide to choose one of the precision. In this case, we consider two different threshold rates. This rate can change based on the maximum required frame rate, required energy for a MAC operation based on the hardware and the total required MAC based on network.   
EHS represents a cutting-edge approach to powering electronic devices without the need for traditional batteries. These systems capture and convert ambient energy from the environment, such as solar light, thermal gradients, vibrations, or radio frequency signals, into electrical power. Once energy is captured, it is stored in a capacitor and then managed and regulated by power management circuits to ensure a stable and usable power supply for the device. By eliminating battery dependency, EHS not only reduces maintenance and replacement costs but also enhances the sustainability and longevity of electronic systems, making them ideal for applications in remote sensing, wearable technology, and IoT. However, a primary challenge for these systems is the instability of the power source. For example, in the case of solar panels, the energy received can fluctuate as a result of varying conditions such as the position of the Sun and cloud cover \cite{roohi2018nv,taheri2024intermittent,tabrizchi2024diac}. 
Consequently, to maximize energy efficiency and device performance over time, these systems must adapt to different energy availability scenarios. To address this issue, we have designed two modules: the Adaptive Scheduler, which comprises the Controller (CTR), and the Smart Power Management (SPM) modules.

These components work together to adjust the operation's precision and expedite computations by utilizing designed early exits in cases of power limitations.
Herein, two critical energy-related factors are considered: \emph{charging rate} ($R_{c}$) and \emph{remaining energy in the capacitor} ($E_{sys}$). To achieve optimal performance, the SPM must continuously monitor both parameters. In the proposed system, EATS, the precision of computation significantly affects both the power consumption rate and the resulting accuracy. Additionally, once computation for a given input begins, the precision level remains constant and cannot be altered mid-process.
Consequently, the SPM measures the charging rate and, based on this measurement, selects an appropriate precision level. In this context, two distinct threshold rates are considered. These thresholds $(R_{\text{th}} )$ 
% can vary depending on factors such as the maximum required frame rate, the energy required for a MAC operation based on the hardware specifications, and the total number of MAC operations required by the network, which 
are determined by the following formula:
\begin{equation}
\label{eq:charge}
R_{\text{th}} = \kappa \times F_{\text{max}} \times N_{\text{MAC}} \times E_{\text{MAC}}
\end{equation}
Where, $R_{\text{th}}$ is charging rate threshold (W),
$\kappa$ is the scaling factor for hardware efficiency and safety margin,
$F_{\text{max}}$ is the maximum required frame rate (fps),
$N_{\text{MAC}}$ is total number of MAC operations per frame, and 
$E_{\text{MAC}}$ is the energy required per MAC operation (J).
Based on the calculated \( R_{\text{th}} \), the SPM module selects the appropriate precision (quantization) level for computations by comparing the current charging rate (\( R_c \)) against predefined thresholds (\( R_{\text{th}_1} \) and \( R_{\text{th}_2} \)). The precision is chosen as follows:
\begin{equation}
\label{eq:thershold}
precision =
\begin{cases} 
\text{32-bit} & \text{if } R_c \geq R_{\text{th}_2} \\
\text{Q8} & \text{if } R_{\text{th}_1} \leq R_c < R_{\text{th}_2} \\
\text{Q4} & \text{if } R_c < R_{\text{th}_1} \\
\end{cases}
\end{equation}
% Another responsibility of SPM is measuring the remaining charge in the system. In this case, as long as the energy of the system is lower than a certain threshold, computation will be done using the nearest early exit. The value of this threshold depends on the energy required for a MAC operation based on the hardware specifications, the maximum number of MAC operations between the start to the first early exit, two early exits and the second early exit with output based on the network architecture.
% Another responsibility of the SPM is measuring the remaining charge in the system. When the system's energy falls below a certain threshold, computations are finished using the nearest early exit. The value of this threshold depends on several factors, including the energy required for a MAC operation based on hardware specifications, the maximum number of MAC operations from the start of computation to the first early exit, from the first early exit to the second one and from the second early exits to the final exit as determined by the network architecture.

% The threshold energy (\( E_{\text{th}} \)) can be calculated using the following formula:

% \[
% E_{\text{th}} = \kappa \times E_{\text{MAC}} \times \left( N_{\text{MAC}_1} + N_{\text{MAC}_2} \right)
% \]

\begin{figure}[b]
    \centering
    \vspace{-0.5em}
    \includegraphics[width=\linewidth]{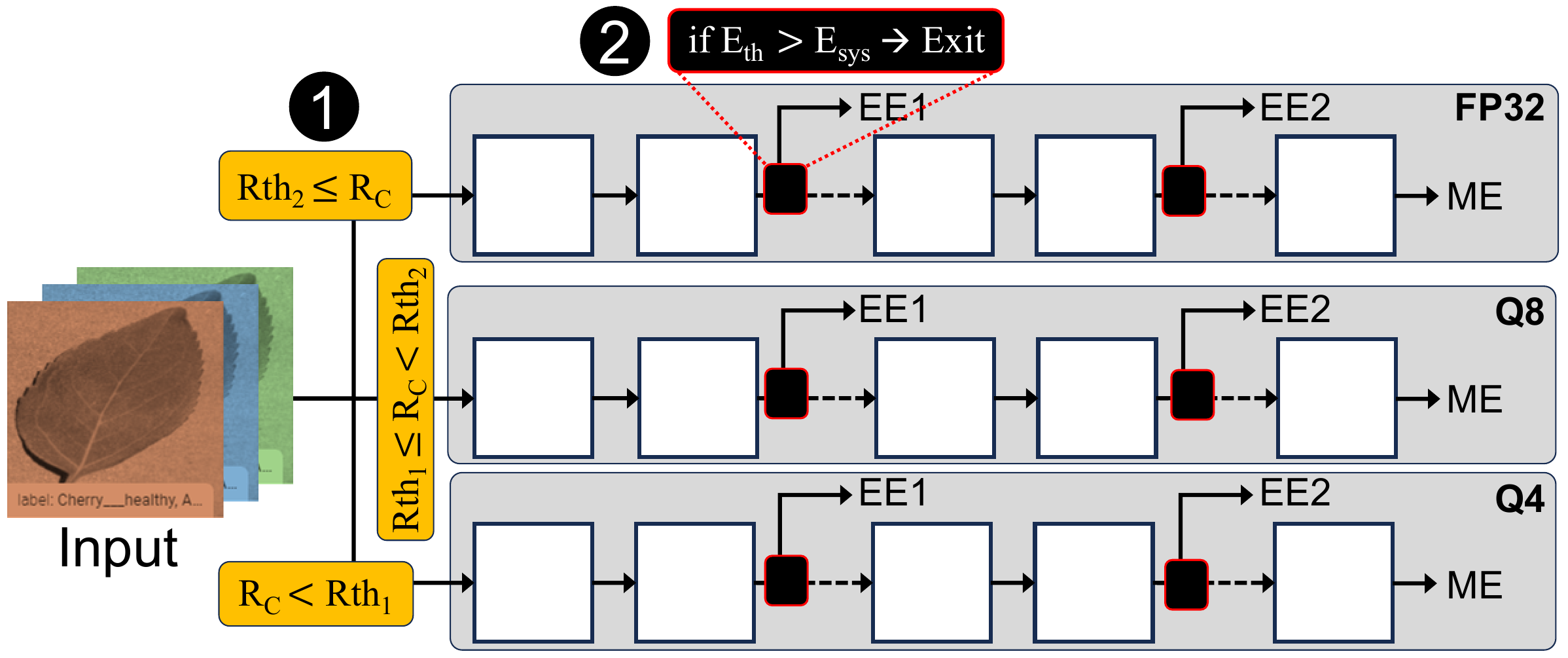}
    \caption{\small Design flow of the proposed ATM-Net architecture.}
    %\vspace{-1em}
    \label{fig:atmnet}
\end{figure}

Another responsibility of the SPM is measuring the remaining charge in the system. When the system's energy falls below a certain threshold, computations are finished using the nearest early exit. The value of this threshold depends on several factors, including the energy required for a MAC operation based on hardware specifications, the maximum number of MAC operations from the start of computation to EE1, from EE1 to EE2, and from EE2 to ME as determined by the network architecture. The threshold energy (\( E_{\text{th}} \)) can be calculated using the following formula:
\begin{equation}
\label{eq:energy}
E_{\text{th}} = \kappa \times E_{\text{MAC}} \times Max\left( N_{\text{MAC}_1} , N_{\text{MAC}_2} , N_{\text{MAC}_3} \right)
\end{equation}
Where,
$E_{\text{th}}$ is the threshold energy (J),
$\kappa$ is the scaling factor for the safety margin,
$E_{\text{MAC}}$ is the energy required per Multiply-Accumulate (MAC) operation (J),
$N_{\text{MAC}_1}$ is the number of MAC from the start to EE1,
$N_{\text{MAC}_2}$ is the number of MAC from EE1 to EE2,
$N_{\text{MAC}_3}$ is the total number of MAC from EE2 to ME.
% The whole structure of ATM-net is shown in Fig.~\ref{fig:atmnet}. In this structure, each quantization level is shown in a different gray box. SPM chooses one of them based on the current charge value $R_{c}$. after that in each exit point ($E$) the energy of system $E_{sys}$ compares to the $E_{th}$. If system has enough energy to continue, the network will meet the next $E$ point. otherwise use that $E$ point to finish the computation. 
% The entire structure of ATM-net is illustrated in Fig.~\ref{fig:atmnet}. In this structure, each quantization level is represented by a distinct gray box. The SPM selects one of these levels based on the current charge value $R_{c}$. Subsequently, at each exit point ($E$), the system's energy $E_{\text{sys}}$ is compared to the threshold $E_{\text{th}}$. If the system has sufficient energy to proceed, the network advances to the next $E$ point. Otherwise, it uses the current $E$ point to terminate the computation. It should mentioned the structure of gray boxes are the same  and we only separate them for better visualization. The only difference between them is the quantization level.
The entire data flow of ATM-net is illustrated in Fig.~\ref{fig:atmnet}. In this structure, each quantization level is represented by a distinct gray box. EATS selects one of these levels based on the current charge value $R_{c}$, step \encircle{1}. 
%Additionally, each gray box shares the same structural design; they are separated only for better visualization. The only difference between them is the quantization level.
Subsequently, at each exit point ($E$), the system's energy $E_{\text{sys}}$ is compared to the threshold $E_{\text{th}}$ in the step \encircle{2}. If the system has sufficient energy to proceed, the network advances to the next $E$ point. Otherwise, it utilizes the current $E$ point to terminate the computation. 
% \neeraj{Please write a sentence with Sep's help, and refer to this figure.}

\section{Results}
\noindent\textbf{Experimental Setup:} 
The analysis of our model parameters, as shown in Table \ref{table:parameters}, MAC operations, and parameter size, highlights the efficiency of our quantization and multi-termination approach. ResNet-18 and DenseNet-121 are evaluated across their full model and various exit points, providing insights into computational demands and memory usage. The MAC operations for the full model are not calculated, as our focus lies on early exit points where computational savings are critical. This omission aligns with the intent of our multi-termination framework, which dynamically halts computation based on input complexity, making full model operations less relevant for such adaptive architectures.
For ResNet-18, the total parameters and MAC operations progressively increase from EE1 (1.58e+05 params, 4.02e+07 MACs) to ME (1.12e+07 params, 1.41e+08 MACs). This trend reflects the hierarchical nature of feature extraction, where deeper layers process increasingly complex representations. Similarly, DenseNet-121 demonstrates a consistent rise in computational complexity, with EE1 (3.79e+05 params, 9.57e+07 MACs) and EE2 (1.30e+06 params, 1.54e+08 MACs) being significantly more efficient compared to ME (1.47e+06 params, 1.62e+08 MACs).
Quantization further optimizes these models, as evident in the parameter size reduction from FP32 to Q8 and Q4. For instance, ResNet-18 at ME sees a drop in parameter size from 47.19 MB (FP32) to 11.28 MB (Q8) and 5.64 MB (Q4). DenseNet-121 exhibits a similar trend, with ME reducing from 5.88 MB (FP32) to 1.47 MB (Q8) and 0.74 MB (Q4). This reduction highlights the resource efficiency achieved without compromising key performance metrics.
The absence of MAC calculations for the full model emphasizes the practical utility of early exits, which align with the energy and latency constraints of edge computing environments. Such efficient resource allocation is critical for deploying these networks in energy-constrained environments, such as edge devices and energy-harvesting systems, while maintaining a balance between accuracy and operational overhead.
\begin{table}[b]
\caption{\small Comparison of model parameters, MAC operations, and parameter sizes across different quantization levels for ResNet-18 and DenseNet-121.}
\resizebox{\columnwidth}{!}{%

\begin{tabular}{llccccc}
\hline
\rowcolor[HTML]{C0C0C0} 
                      &                     & \multicolumn{1}{l}{} & \multicolumn{1}{l}{} & \multicolumn{3}{l}{\textbf{Params Size   (MB)}} \\ \hline
                      \rowcolor[HTML]{C0C0C0} 
\textbf{MODEL} &
  \textbf{Stage} &
  \multicolumn{1}{l}{\textbf{Total Params}} &
  \multicolumn{1}{l}{\textbf{\# MAC}} &
  \multicolumn{1}{l}{\textbf{FP32}} &
  \multicolumn{1}{l}{\textbf{Q8}} &
  \multicolumn{1}{l}{\textbf{Q4}} \\ \hline
\textbf{}             & \textbf{Full Model} & 2.24e+07             & --                   & 47.19          & 21.94          & 10.97         \\
\textbf{RESNET-18}    & \textbf{EE1}     & 1.58e+05             & 4.02e+07             & 1.95           & 0.48           & 0.24          \\
\textbf{}             & \textbf{EE2}     & 6.84e+05             & 7.37e+07             & 4.72           & 1.15           & 0.58          \\
\textbf{}             & \textbf{ME}     & 1.12e+07             & 1.41e+08             & 47.19          & 11.28          & 5.64          \\ \hline
\textbf{}             & \textbf{Full Model} & 8.03e+06             & --                   & 5.91           & 1.48           & 0.74          \\
\textbf{DENSENET-121} & \textbf{EE1}     & 3.79e+05             & 9.57e+07             & 1.52           & 0.38           & 0.19          \\
                      & \textbf{EE2}     & 1.30e+06             & 1.54e+08             & 5.21           & 1.3            & 0.65          \\
                      & \textbf{ME}     & 1.47e+06             & 1.62e+08             & 5.88           & 1.47           & 0.74  \\ \hline       
\end{tabular}
}%\vspace{-2em}
\label{table:parameters}
\end{table}
%%%%\subsubsection{}

\noindent\textbf{Datasets:} 
To evaluate the performance of our models under different quantization schemes, we utilized three diverse datasets: CIFAR-10 \cite{krizhevsky2009learning}, PlantVillage \cite{hughes2015open}, and TissueMNIS \cite{medmnist2021}. CIFAR-10 is a widely used benchmark dataset for image classification tasks, consisting of 60,000 color images divided into 10 classes, with each image having a resolution of $32\times32$ pixels. The dataset is split into 50,000 training images and 10,000 test images, representing objects such as airplanes, birds, and automobiles. Its balanced class distribution and manageable size make it ideal for assessing the general performance of deep learning models.
PlantVillage comprises a large collection of high-resolution images of healthy and diseased plant leaves spanning multiple crop species. For this study, the dataset was preprocessed to standardize input dimensions to $32\times32$ pixels, enabling compatibility with the model architectures. This dataset was chosen to evaluate the models’ capability in real-world agricultural diagnostics, where accurate classification is crucial for early disease detection. 
TissueMNIST is derived from histopathological images of human tissues and consists of grayscale images categorized into nine tissue types. It contains 236,386 images with a standardized resolution of $28\times28$, split into training, validation, and test sets. This dataset is particularly challenging due to its grayscale nature and subtle inter-class variations, making it ideal for testing the robustness of the proposed models in medical image analysis. These datasets collectively provide a comprehensive evaluation framework, encompassing general-purpose classification tasks, domain-specific agricultural diagnostics, and challenging medical imaging applications.

\begin{table}[b]
\vspace{-1em}
\caption{\small Accuracy comparison (in \%) across different quantization schemes and datasets for ResNet-18 and DenseNet-121.}
\resizebox{\columnwidth}{!}{%
\begin{tabular}{l l ccc| ccc| ccc}
\hline
\rowcolor[HTML]{C0C0C0} 
             &             & \multicolumn{3}{c}{\textbf{CIFAR-10}} & \multicolumn{3}{c}{\textbf{PLANTVILLAGE}} & \multicolumn{3}{c}{\textbf{TISSUEMNIST}} \\ \hline
             &             & \textbf{Base} & \textbf{Q8} & \textbf{Q4} & \textbf{Base} & \textbf{Q8} & \textbf{Q4} & \textbf{Base} & \textbf{Q8} & \textbf{Q4} \\ \hline
\multirow{3}{*}{\textbf{RESNET-18}} & \textbf{EE1} & 83.70     & 77.66     & 75.60     & 86.22     & 87.31     & 20.62     & 64.33     & 59.09     & 43.20     \\ 
                                     & \textbf{EE2} & 88.00     & 83.05     & 80.37     & 93.48     & 92.61     & 21.30     & 61.34     & 62.72     & 52.00     \\ 
                                     & \textbf{ME}  & 89.06     & 84.32     & 81.10     & 98.40     & 94.65     & 93        & 62.62     & 59.90     & 57.30     \\ \hline
\multirow{3}{*}{\textbf{DENSENET-121}} & \textbf{EE1} & 84.53     & 78.46     & 73.17     & 92.81     & 91.06     & 90.33     & 50.65     & 43.00     & 41.80     \\ 
                                     & \textbf{EE2} & 88.14     & 82.53     & 80.71     & 96        & 94.90     & 94.39     & 57.95     & 56.47     & 54.33     \\ 
                                     & \textbf{ME}  & 88.35     & 82.57     & 81.18     & 99.81     & 96.93     & 95.43     & 61.00     & 59.52     & 57.59     \\ \hline
\end{tabular}%
}%\vspace{-1.5em}
\label{table:accuracy}
\end{table}

\subsection{Accuracy}
The accuracy results across ResNet-18 and DenseNet-121, illustrated in Table \ref{table:accuracy}, reveal significant insights into the impact of quantization and dataset characteristics on the models’ performance at different exits. For instance, in the PlantVillage dataset, ResNet-18 under Q4 quantization exhibits a stark drop in accuracy at EE2 (21.30\%) and EE1 (20.62\%) compared to its base precision counterparts (93.48\% at EE2 and 86.22\% at EE1). This discrepancy can be attributed to the reduced representational capacity of Q4 quantization, which, when coupled with EEs, struggles to capture the high-resolution features required for accurate classification in PlantVillage—a dataset with a diverse set of high-dimensional images. This highlights the sensitivity of shallow exits to extreme quantization when the dataset demands detailed feature extraction.

Conversely, TissuemNIST presents a unique trend where the base model's EE1 accuracy (64.33\%) surpasses that of the Main exit (62.62\%). This anomaly is likely due to the hierarchical nature of the features in TissuemNIST, where low-level features such as textures and patterns—effectively captured at EE1—are more discriminative for classification than the high-level abstractions processed by deeper layers. This phenomenon underscores the importance of aligning exit placement with dataset-specific feature hierarchies.

\begin{figure}[t]
    \centering
    \includegraphics[width=0.9\linewidth]{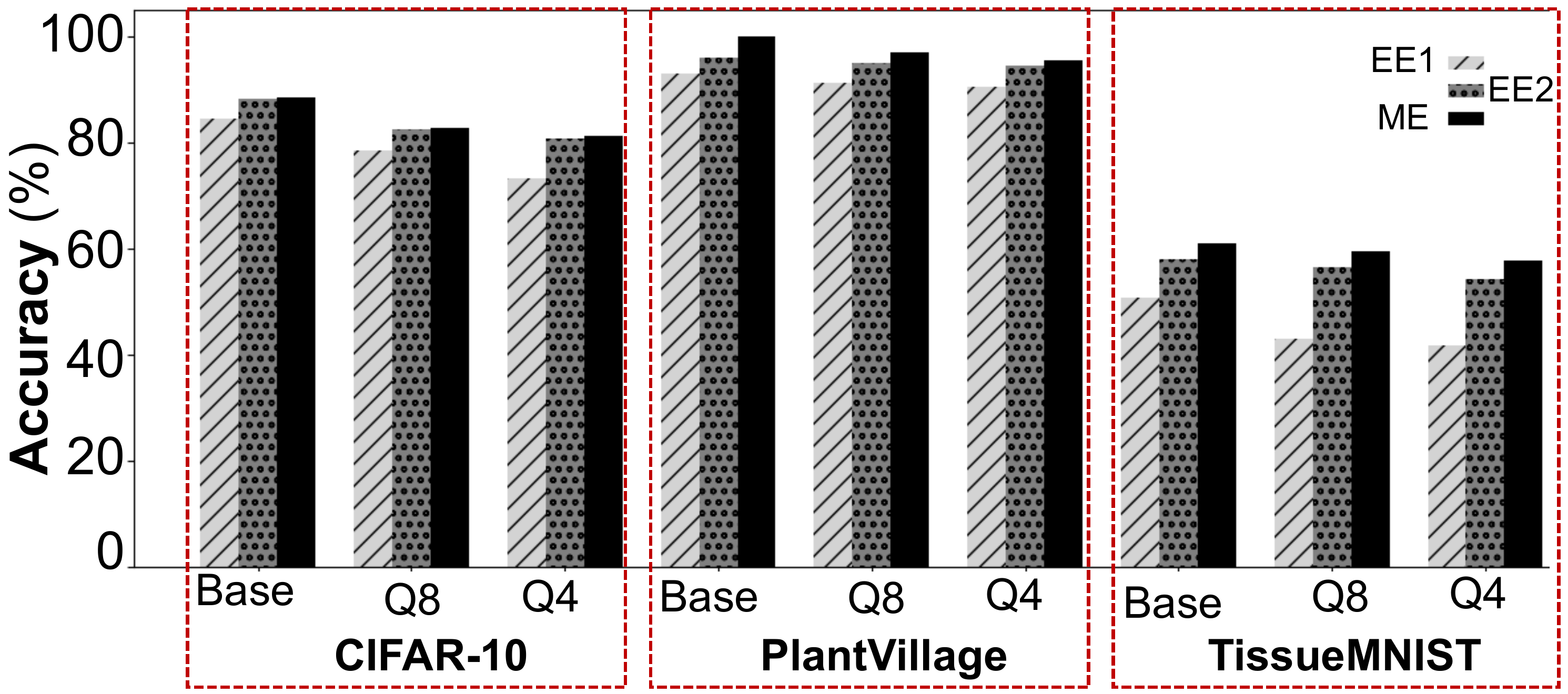}
    \caption{\small Accuracy trends for DenseNet-121 across different datasets.}
    \vspace{-1.5em}
    \label{fig:label_for_image}
\end{figure}

Additionally, DenseNet-121 demonstrates consistent robustness across exits and quantization schemes, particularly in PlantVillage and CIFAR-10, where Q4 accuracy remains above 90\% at all exits. This robustness is evident in Figure~\ref{fig:label_for_image}, which highlights DenseNet-121's ability to maintain high accuracy across varying quantization levels and datasets. This can be attributed to the dense connectivity of DenseNet-121, which facilitates effective feature reuse and compensates for the loss of precision due to quantization. However, in TissuemNIST, DenseNet-121 also suffers a decline in accuracy at deeper exits for Q4 quantization, indicating
that the combined effect of aggressive quantization and complex spatial relationships in the dataset adversely impacts the model’s ability to generalize.

These observations underline the critical interplay between dataset characteristics, model architecture, and quantization levels. While shallow exits under low-precision quantization can provide computational savings, their effectiveness is highly dataset-dependent. This analysis reinforces the need for careful tuning of quantization strategies and exit placements tailored to specific datasets to optimize both accuracy and efficiency.

\subsection{Performance}
To evaluate the performance of the proposed ATM-Net, the examined networks were implemented on the Xilinx Artix-7 XC7A200T FPGA, leveraging its DSP48E1 slices. This FPGA features 740 DSP slices, each capable of high-performance arithmetic operations essential for MAC computations, which are foundational in deep learning inference. Additionally, the FPGA includes 215,360 logic cells and 13,140 Kb of block RAM, providing the resources required for computationally intensive workloads while maintaining energy efficiency.

The implementation targeted three precision levels: FP32, Q8, and Q4, as shown in Table \ref{table:energy}. The DSP48E1 slices, originally designed for high-precision arithmetic, were reconfigured to support lower-precision operations by optimizing their internal structures. This approach allowed efficient use of hardware resources and reduced both power consumption and computational delay. For Q8 and Q4 operations, a similar technique was employed to map multiple smaller precision inputs onto the DSP slices. The operands were packed into the 48-bit accumulator and $25\times18$ multiplier of the DSP slice. For Q8 operations, each DSP slice processed multiple MAC operations in parallel by packing the Q8 inputs into the wider input and accumulator fields, thereby leveraging the full hardware capacity. Similarly, for Q4 operations, the inputs were further packed to allow even greater parallelism, with each DSP slice supporting up to 12 MAC operations concurrently. This method maximized computational throughput and minimized resource utilization. The reduced bit widths also lowered the switching activity within the DSP slice, contributing to substantial reductions in power consumption and delay. At FP32 precision, the DSP slices were used to their full capacity, performing a single MAC operation per slice due to the higher bit-width requirements of floating-point arithmetic. While this provided high computational accuracy, it came at the cost of higher power consumption and delay.

\begin{table}[h]
\caption{\small The Results of power, delay, and PDP for different networks and precisions.}
\resizebox{\columnwidth}{!}{
\begin{tabular}{lccc|ccc}
\hline 
\rowcolor[HTML]{C0C0C0} 
                                & \multicolumn{3}{c}{\textbf{RESNET-18}}                           & \multicolumn{3}{c}{\textbf{DENSENET121}}                         \\
                                \hline
                                &  \textbf{Power} & \textbf{Delay} & \textbf{PDP}  & \textbf{Power} & \textbf{Delay} & \textbf{PDP}  \\
\hline
\textbf{EE1 (FP32)}  & 9.36e+00       & 8.92e-02       & 8.34e-01     & 2.23e+01       & 2.13e-01       & 4.74e+00     \\
\textbf{EE2 (FP32)}  & 1.72e+01       & 1.64e-01       & 2.81e+00     & 3.59e+01       & 3.42e-01       & 1.23e+01     \\
\textbf{ME (FP32)}  & 3.28e+01       & 3.13e-01       & 1.03e+01     & 3.78e+01       & 3.60e-01       & 1.36e+01     \\ \hline
\textbf{EE1 (Q8)} & 2.17e+00       & 1.49e-02       & 3.22e-02     & 5.17e+00       & 3.54e-02       & 1.83e-01     \\
\textbf{EE2 (Q8)} & 3.98e+00       & 2.73e-02       & 1.09e-01     & 8.31e+00       & 5.70e-02       & 4.74e-01     \\
\textbf{ME (Q8)} & 7.61e+00       & 5.21e-02       & 3.96e-01     & 8.77e+00       & 6.01e-02       & 5.27e-01     \\ \hline
\textbf{EE1 (Q4)} & 1.16e+00       & 7.43e-03       & 8.65e-03     & 2.78e+00       & 1.77e-02       & 4.92e-02     \\
\textbf{EE2 (Q4)} & 2.14e+00       & 1.36e-02       & 2.91e-02     & 4.46e+00       & 2.85e-02       & 1.27e-01     \\
\textbf{ME (Q4)} & 4.08e+00       & 2.61e-02       & 1.06e-01     & 4.71E+00       & 3.00e-02       & 1.41e-01       \\ \hline
\end{tabular}
\vspace{-2em}
\label{table:energy}
}
\end{table}

The experimental results in Table \ref{table:energy} highlight the efficiency of lower precision arithmetic. For example, in the final output stage of DenseNet-121, power consumption decreased from 37.8 W for FP32 to 8.77 W and 4.71 W for Q8 and Q4 precisions, respectively. Similarly, the delay improved from 0.36 ms for FP32 to 0.06 ms for Q8 and 0.03 ms for Q4. The power-delay product (PDP) followed a similar trend, improving from 13.6 J for FP32 to 0.527 J (Q8) and 0.141 J (Q4). For ResNet-18, similar improvements were observed; the PDP for the final output stage decreased from 10.3 J for FP32 to 0.396 J and 0.106 J for Q8 and Q4, respectively. 
These results demonstrate the significant energy efficiency and computational benefits of adopting lower precision arithmetic while leveraging the proposed architecture. The observed trends validate the effect of using lower bit widths for MAC operations in battery-less systems where energy efficiency is a critical requirement.

\begin{figure}[h]
    \centering
    \vspace{-1em}
    \includegraphics[width=\linewidth]{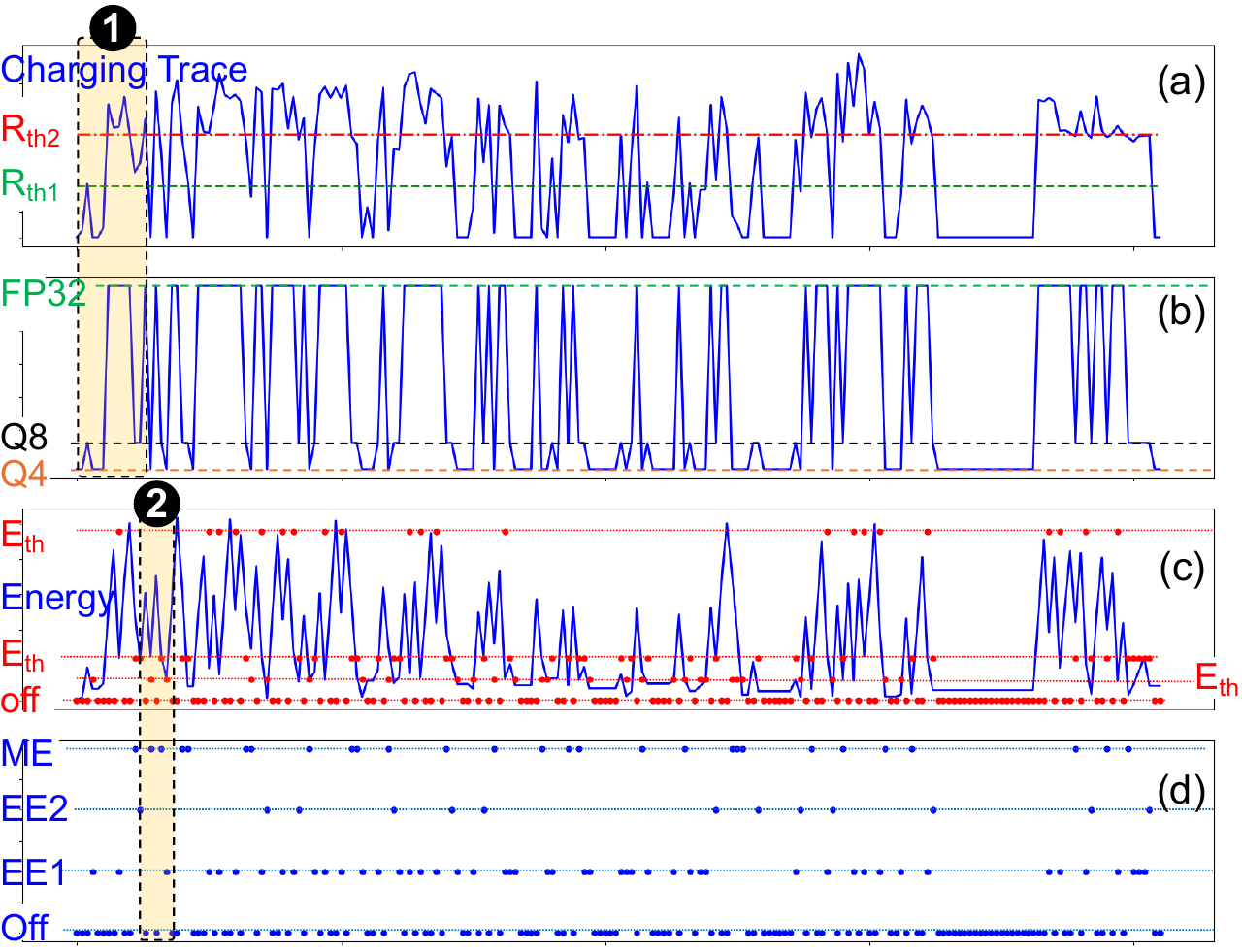}
    \caption{\small (a) The charging rate. The (b) quantization level, (c) system's energy at runtime, and (d) behavior of the system w.r.t exit points.}
    \vspace{-1em}
    \label{fig:sim}
\end{figure}

To validate the energy-aware task scheduler (EATS) module, we conduct system simulations using actual charging rate data. The results are presented in Fig.~\ref{fig:sim}. 
Figure~\ref{fig:sim}(a) illustrates the charging rate along with two thresholds, $R_{\text{th1}}$ and $R_{\text{th2}}$. Figure~\ref{fig:sim}(b) shows the precision levels of the operations, as defined in Eqs.~\ref{eq:charge} and \ref{eq:thershold}. For instance, in \encircle{1}, when the charging rate exceeds $R_{\text{th1}}$ but remains below $R_{\text{th2}}$, the precision transitions from Q4 to Q8. Subsequently, when the charging rate surpasses $R_{\text{th2}}$, the precision increases to FP32.
Figure~\ref{fig:sim}(c) depicts the system's energy alongside the energy threshold, $E_{\text{th}}$. As described in Eq.~\ref{eq:energy}, $E_{\text{th}}$ corresponds to the energy consumption per MAC operation and varies across precision levels (FP32, Q8, and Q4) due to their distinct energy requirements. When $E_{\text{th}}$ reaches zero, it indicates that the system has depleted its energy and is unable to continue executing the network.
Figure~\ref{fig:sim}(d) represents the system's early exits. In this figure, the system waits until it accumulates sufficient energy to exceed $E_{\text{th}}$ before beginning computation. At each exit point (EE1, EE2, and ME), EATS evaluates the remaining energy to decide whether to continue computation or use the nearest early exit based on the current energy state. For example, in \encircle{2}, the system starts in an off state but eventually accumulates enough energy to initiate computation. At this point, the system completes the entire computation and exits at the ME point twice. In the subsequent start phase, the system's energy depletes quickly, leading it to perform computations through EE1.

%%%\sepehr{I think we need some results using one power trace and applying your CTR and SPM}

\section{Conclusion}
ATM-Net offers a robust framework for deploying deep neural networks on energy-harvested IoT devices. Integrating multi-precision computing with adaptive early exits addresses power instability in energy harvesting systems. Experimental results highlight significant energy savings, with DenseNet-121 achieving over 90\% accuracy at Q4 precision while reducing power consumption by up to 87.5\%. The energy-aware scheduler effectively manages the precision-accuracy trade-off, establishing ATM-Net as a promising solution for energy-efficient, batteryless edge intelligence. Future research could explore dynamic precision adjustment and dataset-specific optimization of exit points.

\bibliographystyle{IEEEtran}

\bibliography{IEEEabrv,./main}\vspace{-2em}

\begin{thebibliography}{10}
\providecommand{\url}[1]{#1}
\csname url@samestyle\endcsname
\providecommand{\newblock}{\relax}
\providecommand{\bibinfo}[2]{#2}
\providecommand{\BIBentrySTDinterwordspacing}{\spaceskip=0pt\relax}
\providecommand{\BIBentryALTinterwordstretchfactor}{4}
\providecommand{\BIBentryALTinterwordspacing}{\spaceskip=\fontdimen2\font plus
\BIBentryALTinterwordstretchfactor\fontdimen3\font minus \fontdimen4\font\relax}
\providecommand{\BIBforeignlanguage}[2]{{%
\expandafter\ifx\csname l@#1\endcsname\relax
\typeout{** WARNING: IEEEtran.bst: No hyphenation pattern has been}%
\typeout{** loaded for the language `#1'. Using the pattern for}%
\typeout{** the default language instead.}%
\else
\language=\csname l@#1\endcsname
\fi
#2}}
\providecommand{\BIBdecl}{\relax}
\BIBdecl

\bibitem{verma2022paradigm}
P.~Verma \emph{et~al.}, ``A paradigm shift in iot cyber-security: A systematic review,'' \emph{International Journal for Research in Applied Science and Engineering Technology}, vol.~10, no.~10, pp. 1024--1034, 2022.

\bibitem{liu2020energy}
S.~Liu \emph{et~al.}, ``Energy-aware mac protocol for data differentiated services in sensor-cloud computing,'' \emph{Journal of cloud computing}, vol.~9, pp. 1--33, 2020.

\bibitem{ericsson2024climate}
{Ericsson}, ``Climate action - ericsson,'' \url{https://www.ericsson.com/en/about-us/sustainability-and-corporate-responsibility/environment/climate-action}, 2024, accessed: 2024-05-01.

\bibitem{cordis2024batteries}
\BIBentryALTinterwordspacing
{European Commission}. (2024) Up to 78 million batteries will be discarded daily by 2025, researchers warn. Accessed: 2024-05-01. [Online]. Available: \url{https://cordis.europa.eu/article/id/430457-up-to-78-million-batteries-will-be-discarded-daily-by-2025-researchers-warn}
\BIBentrySTDinterwordspacing

\bibitem{teerapittayanon2016branchynet}
S.~Teerapittayanon \emph{et~al.}, ``Branchynet: Fast inference via early exiting from deep neural networks,'' in \emph{2016 23rd international conference on pattern recognition (ICPR)}.\hskip 1em plus 0.5em minus 0.4em\relax IEEE, 2016, pp. 2464--2469.

\bibitem{jacob2018quantization}
B.~Jacob \emph{et~al.}, ``Quantization and training of neural networks for efficient integer-arithmetic-only inference,'' in \emph{Proceedings of the IEEE Conference on Computer Vision and Pattern Recognition (CVPR)}, 2018, pp. 2704--2713.

\bibitem{resnet2015}
K.~He \emph{et~al.}, ``Deep residual learning for image recognition,'' in \emph{Proceedings of the IEEE Conference on Computer Vision and Pattern Recognition (CVPR)}, 2016, pp. 770--778.

\bibitem{densenet2017}
G.~Huang \emph{et~al.}, ``Densely connected convolutional networks,'' in \emph{Proceedings of the IEEE Conference on Computer Vision and Pattern Recognition (CVPR)}, 2017, pp. 4700--4708.

\bibitem{roohi2018nv}
A.~Roohi and R.~F. DeMara, ``Nv-clustering: Normally-off computing using non-volatile datapaths,'' \emph{IEEE Transactions on Computers}, vol.~67, no.~7, pp. 949--959, 2018.

\bibitem{taheri2024intermittent}
N.~Taheri \emph{et~al.}, ``Intermittent-aware design exploration of systolic array using various non-volatile memory: A comparative study,'' \emph{Micromachines}, vol.~15, no.~3, p. 343, 2024.

\bibitem{tabrizchi2024diac}
S.~Tabrizchi \emph{et~al.}, ``Diac: Design exploration of intermittent-aware computing realizing batteryless systems,'' in \emph{2024 Design, Automation \& Test in Europe Conference \& Exhibition (DATE)}.\hskip 1em plus 0.5em minus 0.4em\relax IEEE, 2024, pp. 1--6.

\bibitem{krizhevsky2009learning}
\BIBentryALTinterwordspacing
A.~Krizhevsky and G.~Hinton, ``Learning multiple layers of features from tiny images,'' University of Toronto, Tech. Rep., 2009. [Online]. Available: \url{https://www.cs.toronto.edu/~kriz/learning-features-2009-TR.pdf}
\BIBentrySTDinterwordspacing

\bibitem{hughes2015open}
D.~P. Hughes and M.~Salathé, ``An open access repository of images on plant health to enable the development of mobile disease diagnostics,'' \emph{arXiv preprint arXiv:1511.08060}, 2015.

\bibitem{medmnist2021}
J.~Yang \emph{et~al.}, ``Medmnist v2: A large-scale lightweight benchmark for 2d and 3d biomedical image classification,'' \emph{arXiv preprint arXiv:2110.14795}, 2021.

\end{thebibliography}

\end{document}